\ifdefined\pdfinclusioncopyfonts
\fi

\documentclass[
]{ceurart}

\usepackage{listings}
\usepackage{comment}
\usepackage{multirow}
\begin{document}

\copyrightyear{2026}
\copyrightclause{Copyright for this paper by its authors.
  Use permitted under Creative Commons License Attribution 4.0
  International (CC BY 4.0).}

\conference{}

\title{Automated Detection and Encoding of Cartographic Signs in Historical Maps}

\author[1,2]{Remi Petitpierre}[%
orcid=0000-0001-9138-6727,
email=remi.petitpierre@epfl.ch,
]
\address[1]{EPFL, Swiss Federal Institute of Technology in Lausanne. Digital Humanities Laboratory (DHLAB), School of Computer and Communication Sciences (IC). CM 1 468, Station 10, 1015 Lausanne, Switzerland.}
\address[2]{EPFL. Laboratory of Urban Sociology (LASUR), School of Architecture, Civil and Environmental Engineering (ENAC).}

\begin{abstract}
  Symbols and icons in archival maps represent an expansive source of evidence for historical geography. In this paper, we present a two-stage framework for (1) detecting cartographic signs in historical maps and (2) generating interpretable sign embeddings that capture both semantics and stylistic variations. First, we introduce CaSiDD, an open benchmark dataset spanning multiple sign categories and diverse cartographic styles, designed for generic sign detection. Using YOLOv10-X, we demonstrate the practicality of automated sign detection on historical maps. We then present strategies for creating contrasted, informative embeddings of cartographic signs. We approach embedding design systematically, maximizing expressiveness while suppressing digitization or background material confounders. Our final method, based on an adapted RADIOv4 backbone, effectively encodes both sign form and semantics, while producing interpretable representations. This work lays a foundation for scalable evidence extraction for the comparative study of map signs within cartographic culture and geohistorical research.
\end{abstract}


\maketitle

\section {Introduction}

In this work, we present methods to detect and encode cartographic signs in historical map images. More precisely, we focus on \enquote{point signs}, that is, geographic features depicted as discrete, extensionless cartographic objects \cite{bertin_semiologie_1967}. Cartographic signs, such as icons and symbols, are widespread in historical and contemporary maps. For Dainville, they are an integral part of the \emph{language of geographers} \cite{dainville_langage_1964}. Their usage includes, for instance, the representation of natural features (trees, wetlands, hills), settlements, and infrastructure (bridges, gates, mills, train stations). Cartographic signs thus convey information that can be valuable for historical geography, urban economic history, and environmental research.  In this context, the potential for sign detection from historical maps is perceived as an instrument for comparative research rather than for exhaustive reconstruction of past landscapes. Example applications of automated sign extraction could include studying the historical prevalence of tree taxa (e.g. palm trees, broadleaf, coniferous), the economic intensity of river use (e.g. watermills), or the presence of archaeological remains (e.g. chapels, graves). The detection and visual encoding of signs at a large scale may also open new opportunities for the quantitative history of cartography, for semiotic research, and for the cultural geography of maps. Indeed, cartographic signs contain two concurrent evidentiary layers. On the one hand, they provide information about past geography. On the other hand, they also document representational conventions, workshop practices, or a certain cartographic \enquote{grammar}.

For both objectives, the computational process involves two successive operations: detecting map signs and encoding them. The first part of this article therefore assesses the practicality of automated point-sign extraction from historical maps, with the particular aim of developing detection techniques robust to both content and form diversity. Accordingly, we introduce CaSiDD, the Cartographic Sign Detection Dataset, a resource comprising 18,750 sign labels from 796 distinct maps. CaSiDD is conceived as a reusable dataset for pretraining, supporting research-driven extraction tasks, and benchmarking technical improvements. CaSiDD is an indispensable component for addressing signs at scale, across collections, and in their multiplicity.

The second part of the article evaluates encoding approaches and the pragmatic utility of sign embeddings for downstream classification or inference tasks. The ambition in this respect is to develop domain-adapted, interpretable encoders with strong semantic and stylistic expressiveness. We pursue a task-agnostic approach and deliberately enforce invariances to suppress material and collection-induced confounders. These objectives are pursued through targeted experiments testing classification and style recognition capabilities, e.g. workshop attribution. The interpretability of sign embeddings is assessed qualitatively through visualization and empirically through their statistical association with intuitive visual variables representing color and shape.

\section{Related work}

\subsection{Detection of cartographic signs}

Despite the ubiquity of cartographic signs, and the wealth of information they contain, few studies have attempted to extract icons and symbols from maps. By contrast, a growing number of publications have focused on the reputedly more complex task of text spotting \cite{chiang_recognizing_2015,kim_mapkurator_2023,weinman_deep_2019,zou_recognizing_2025,li_icdar_2024}. The detection of signs from maps involves specific challenges that disrupt standard object detection paradigms, including their small size, ambiguous boundaries, dense layouts, and frequent overlaps. To the best of our knowledge, the first article describing the application of modern techniques to map icons was published by Smith et al. \cite{smith_estimating_2025}, who extracted urban trees from the Ordnance Survey map of Leeds. In their study, Smith et al. adopted Mask R-CNN \cite{he_mask_2018}, a classical region-based convolutional neural network. A distinct project focused on extracting icons from the Cassini map of France \cite{sutty_crh-ehesscocass-benchmarking_2024}. In that work, Sutty and Dum\'enieu selected YOLOv10-X \cite{wang_yolov10_2024} after comparing its performance against DETR \cite{carion_end--end_2020} and Detectron2 \cite{wu_detectron2_2019}. Recent YOLO models have consistently outperformed Mask R-CNN across varied benchmarks \cite{sapkota_comparing_2024,wang_yolov7_2023}. YOLO (You Only Look Once) offers computationally efficient object detection models with competitive accuracy \cite{redmon_you_2016}. YOLOv10 is a recent iteration of this family. Compared to its predecessors, it offers greater efficiency with comparable accuracy. It relies on dual-assignment consistency to resolve multiple superimposed detections, thereby obviating the need for non-maximum suppression (NMS) in post-processing, which had hitherto constituted a major computational bottleneck.

Previous work, therefore, did not address the detection of cartographic signs from a generic, cross-collection perspective, nor did it explore the possibility of studying signs from a quantitative, semiotic perspective through visual latent representations. The present study, by contrast, engages directly with the issue of semiotic ambiguity, which is usually outside the scope of object detection benchmarks.

\subsection{Image encoding and embedding}
The objective of image encoding is to generate expressive mathematical representations of graphical content and style for comparison, retrieval, classification, or other downstream tasks. Historically, visual features were extracted using algorithms such as the Histogram of Oriented Gradients (HOG, \cite{mcconnell_method_1986}) and the Scale-Invariant Feature Transform (SIFT, \cite{lowe_object_1999,ruf_mobile_2010,seguin_visual_2016}). Local descriptors were typically aggregated into global representations, or visual Bag-of-Words models \cite{csurka_visual_2004,jegou_aggregating_2010,perronnin_fisher_2007}. Since the 2010s, deep encoders trained via unsupervised or semi-supervised methods have become prominent \cite{mienye_deep_2025}. Variational Autoencoders (VAEs) extended the neural paradigm by adopting a Bayesian approach, resulting in probabilistic latent spaces \cite{kingma_auto-encoding_2022}. Vector-Quantized Variational Autoencoders (VQ-VAE, \cite{oord_neural_2017}) encode images into discrete, learned vocabularies, effectively reducing image tokens to visual prototypes. Vision Transformers (ViT, \cite{dosovitskiy_image_2021}), by contrast, can explicitly integrate multiple local features into a unified global representation, producing a holistic representation of the image sign.

Vision foundation models (VFMs), such as CLIP \cite{radford_learning_2021}, SimCLR \cite{chen_simple_2020}, Stable Diffusion \cite{rombach_high_2022}, MAEs \cite{he_masked_2021}, and DINO \cite{caron_emerging_2021}, often build on the ViT architecture, capitalizing on large image databases or multimodal datasets for pretraining. This confers on them generic capabilities for encoding images into latent representations and semantic spaces. Some works have specifically attempted to adapt VFMs to historical maps. Xia et al. presented a contrastive ViT pretraining approach for railway detection \cite{xia_contrastive_2023}. They also explored the potential of adapting SAM \cite{kirillov_segment_2023} models for the segmentation of geographic features from map series \cite{xia_mapsam_2025,xia_mapsam2_2025}. Sterzinger et al. \cite{sterzinger_fewshot_2026} benchmarked several encoder models, including SAM, DINO, and RADIO, demonstrating the strong performance of the latter on generic segmentation tasks. RADIO \cite{heinrich_radiov25_2025} is an agglomerative vision encoder that distills learned representations from distinct VFMs, including CLIP \cite{radford_learning_2021}, SigLIP2 \cite{tschannen_siglip_2025}, DINO \cite{oquab_dinov2_2024,simeoni2025dinov3}, and Segment Anything (SAM, \cite{kirillov_segment_2023}), into a single model. It aims to produce task-agnostic embeddings with robust generalization capabilities. C-RADIOv3 and C-RADIOv4, its most recent iterations, have consistently demonstrated state-of-the-art performance across various benchmarks, including image-level representation and pixel-level segmentation \cite{radiov3_github,ranzinger_radiov4_2026}.

\section{Automated detection of map signs}

\subsection{Dataset}

\begin{figure}
\centering
\includegraphics[width=\linewidth]{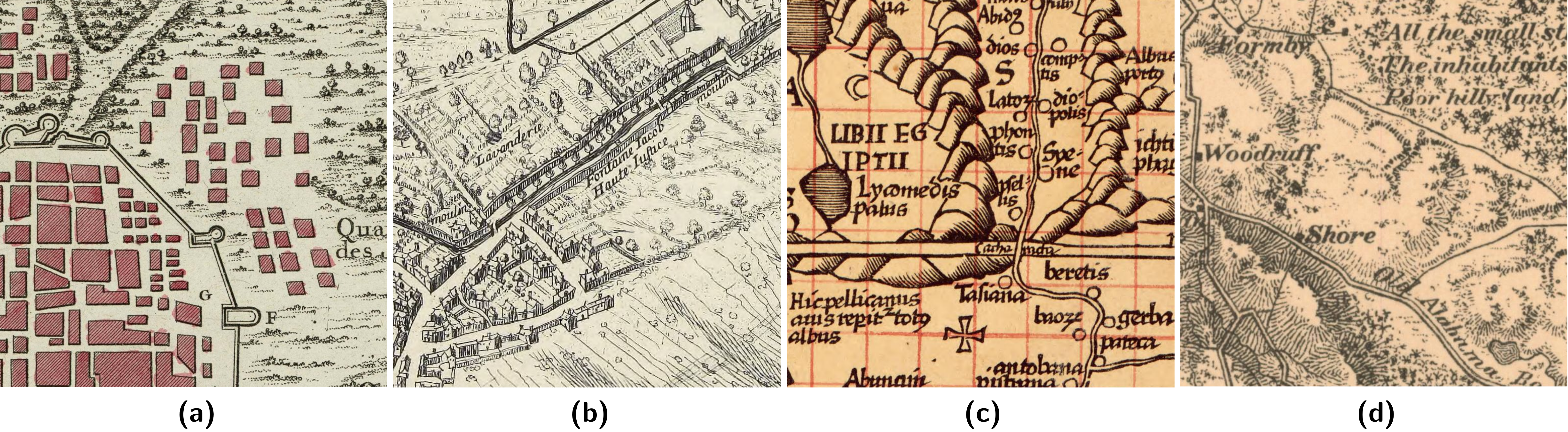}
\caption{Illustration of the challenges to the computational assumption of sign discreteness and the pictoriality of the representation. Signs may overlap (e.g. Fig. \ref{fig:iconicity}c), exhibit ill-defined, ambiguous boundaries (e.g. Fig. \ref{fig:iconicity}a, \ref{fig:iconicity}d), or appear utterly pictorial (e.g. Fig. \ref{fig:iconicity}b).}
\label{fig:iconicity}
\end{figure}

The first step in achieving a generic detection model involves the curation of a broad and diverse training dataset. A new dataset was therefore compiled and manually labeled for the purpose of this study. The Cartographic Sign Detection Dataset (CaSiDD) comprises 796 map patches from a distinct maps, with dimensions ranging from $768 \times 768$ to $1000 \times 1000$ pixels. The dataset is published open access along with this paper \cite{petitpierre_cartographic_2025}. The map patches in CaSiDD are sampled from Semap, a diverse dataset for map semantic segmentation \cite{petitpierre_semantic_2025}, aggregating images from over 20 distinct digital library portals in five countries. Representative image samples are provided in Figure \ref{fig:iconicity}.

In total, 18,750 icons and symbols were annotated. Annotation was performed in a class-agnostic fashion; that is, it was not a priori restricted to a closed ontology of geographic-cartographic objects. This design choice aims to broaden the scope of the dataset and that of the sign detection model by engaging with the diversity of map symbols. This first allows the semiotic continuum to be preserved and later categorized within a smooth, interpretable visual representation space.

Experiment 1a (\ref{subsection:exp1a}) and the multiclass object detection trial assess the ability to classify these signs into distinct cartographic-geographic categories. For this specific purpose, signs were sorted, whenever possible, into 24 frequent categories. Although certain classes of signs, such as those signifying \emph{trees} or \emph{settlements}, appear very frequently across maps, others are much rarer, such as \emph{embassy}, \emph{battlefield}, \emph{lighthouse}, or \emph{post office}. The annotation statistics suggest that icon categorical frequencies tend to follow a power-law distribution overall\footnote{The likelihood ratio of the power-law distribution against exponential is 137 ($pval = 3 \cdot 10^{-7}$).}. This presages difficulty in the categorization task due to class imbalance.

A difficulty of the annotation task is the ill-defined spatial boundedness of icons or symbols. Although automatic detection methods prescribe strictly discrete bounding boxes, signs do not always conform to them straightforwardly. Figure \ref{fig:iconicity} presents challenging examples, showing that the spatial extent of signs is often subjective. In Figure \ref{fig:iconicity}a, for instance, the distinctions among vegetation classes (trees, bushes, grasses) and between these classes and background texture inherently depend on personal interpretation. In Figures \ref{fig:iconicity}b and \ref{fig:iconicity}d, some trees—symbolized in the latter by a star icon—are clearly delineated, while others coalesce into an indistinct tangle. In Figure \ref{fig:iconicity}b, buildings likewise tend to overlap. Finally, Figure \ref{fig:iconicity}c illustrates the challenges of occlusion for hill icons. Such occlusion directly interferes with the assumption of non-superimposition often adopted in object detection algorithms.

\subsection{Training}

The YOLOv10-X model, pretrained on the COCO dataset \cite{lin_coco_2014}, was fine-tuned to convergence using an AdamW optimizer and a decaying learning-rate schedule, initialized at $4 \cdot 10^{-4}$. The training data were augmented through random cropping to $640 \times 640$ pixels, horizontal flipping, and mild jittering of color value ($\pm 8 \%$) and saturation ($\pm 14 \%$). Although common in the YOLO framework, mosaic augmentation was disabled given the already small size of target objects and the risk of disrupting cartographic context.

\begin{figure}
\centering
\includegraphics[width=\linewidth]{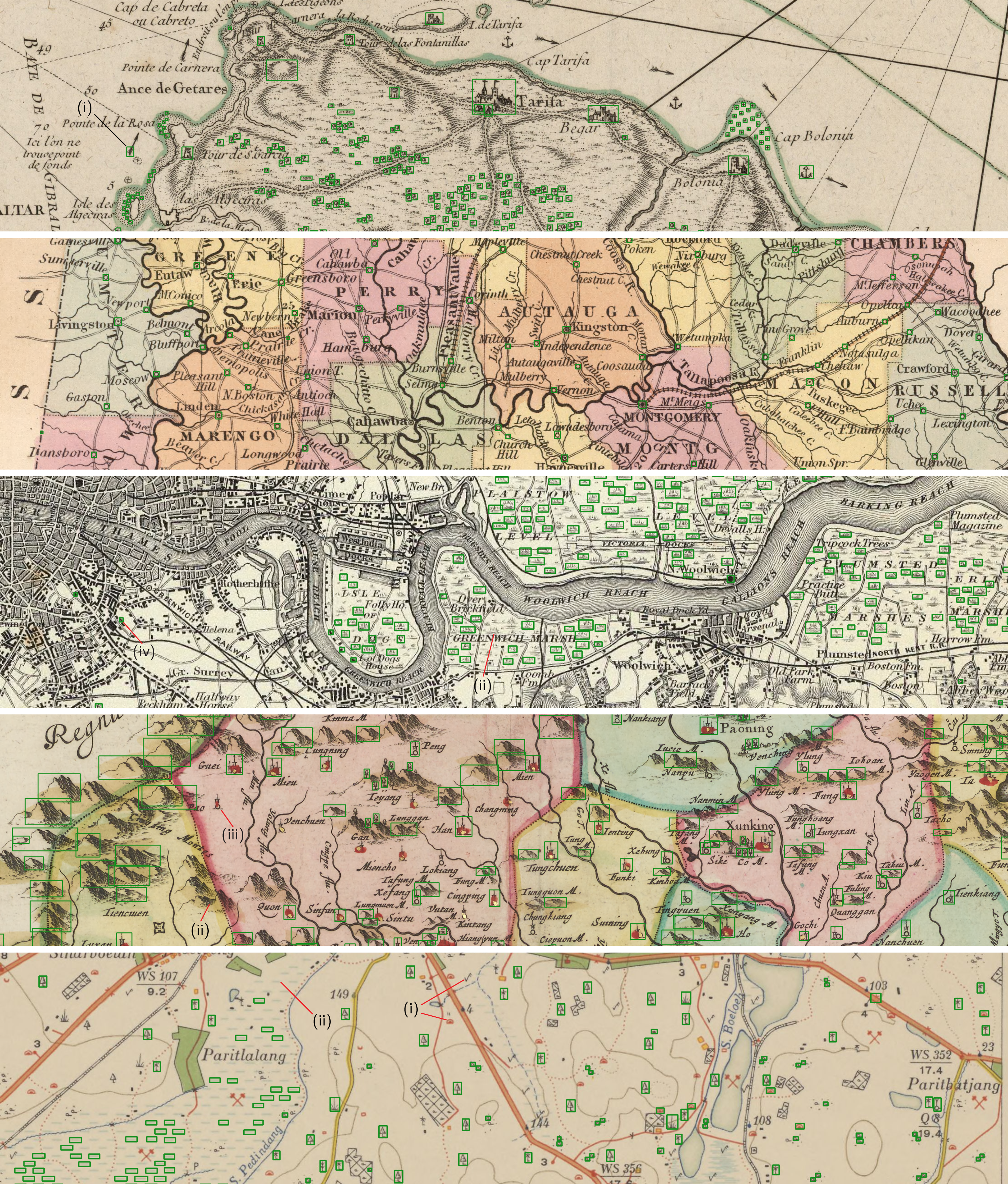}
\caption{Examples of generalization of detection performance outside the dataset. The red lines highlight characteristic error types: (i) atypical occurrences; (ii) unclear boundaries or texture patterns; (iii) occlusion; (iv) small polygons (false positives).}
\label{fig:detection_results}
\end{figure}

\subsection{Results and Assessment of Detection}

Detection performance was validated quantitatively by computing precision and recall at defined intersection-over-union (IoU) thresholds. Performance decreases as this threshold increases. At the baseline level (IoU $\geq 30 \%$), precision reached $81.1 \%$ and recall $41.5 \%$. When the requirement was raised to IoU $\geq 50 \%$, precision was $75.7 \%$ and recall was $38.9 \%$. This sensitivity to the IoU threshold can be caused by loose predicted bounding boxes.

In the multiclass trial ($n_{\text{classes}} =24$), the reported AP50 is $44.6 \%$ and recall is $21.8 \%$. The low recall is probably caused by the steep class imbalance, which compounds semantic ambiguities and sign localization challenges. This result justifies treating categorization as a downstream task.

Figure \ref{fig:detection_results} shows representative examples of single-class detection. Qualitatively, detection results are satisfactory. In the illustrated examples, only a few false positives are visible, and false negatives appear relatively rare. Failure to detect seems more likely for underrepresented sign occurrences (case i), when discreteness is debatable (case ii), or under occlusion (case iii). False positives are often related to the presence of tiny areal objects, such as small buildings, confused with dimensionless signs (case iv). Overall, errors often appear to stem from interpretative divergences rather than solely from technical failures.

\section{Encoding sign style and semantics}

We consider two downstream use cases for the extracted signs: semantic classification and stylometry. In the classification task, we examine the possibility of categorizing signs into distinct classes of geographic objects, like \emph{trees}, \emph{settlements}, or \emph{mills}. In the stylometric task, we assess the potential for retrieving the mapmakers' communities documented in the metadata based on iconographic style and recurrent sign forms.

\subsection{Approach}

Aiming for genericity, we implement a unified, task-agnostic procedure for evaluation in both use cases. The approach relies on adapted VFM backbones to generate embeddings. Specifically, we evaluate four distinct models: DINOv2-ViT-L (304 M parameters), DINOv3-ViT-L (303 M parameters), C-RADIOv3-ViT-B (98 M parameters) and C-RADIOv4-ViT-H (652 M parameters)\footnote{For C-RADIOv3, a lower performance was observed for C-RADIOv3-ViT-L compared to C-RADIOv3-ViT-B. For C-RADIOv4, ViT-H was the smallest model size available at the time of writing.}.

\begin{figure}[b!]
\centering
\includegraphics[width=\linewidth]{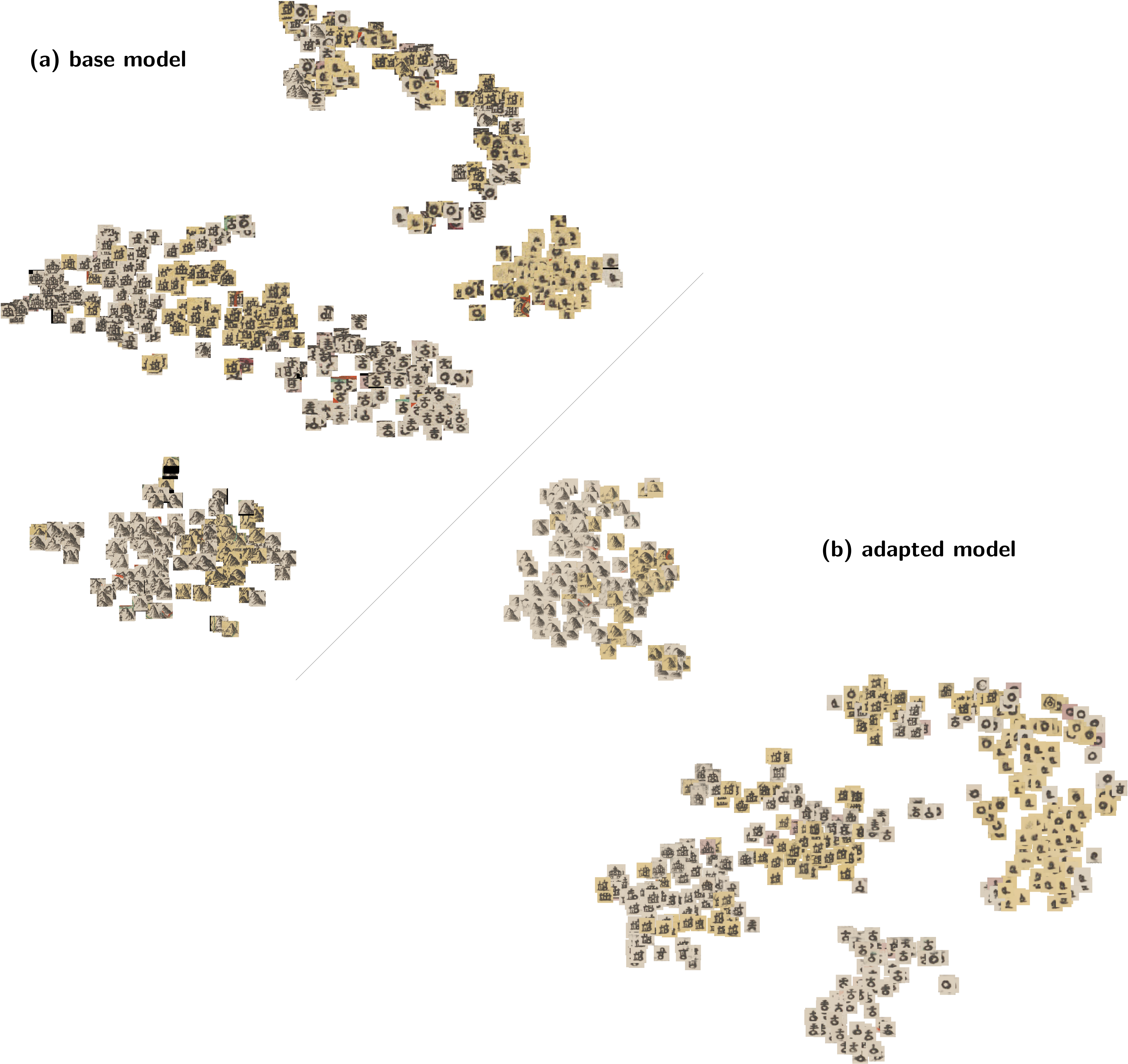}
\caption{Two-dimensional, constrained t-SNE visualization of the sign embeddings, derived from an iconographic map of the Canton of Zurich, ca. 1770 \cite{walser_zurich_1770}. \textbf{(a)} In the base model using RADIOv4, spurious clusters appear (e.g. at the top), and background color heavily affects spatial structure. \textbf{(b)} In the adapted model, clustering is more shape-dependent and distinctive.}
\label{fig:embeddings}
\end{figure}

The main challenges in the present context are (1) the occurrence of spurious or encroaching visual objects, which can introduce noise into the latent representation of cartographic signs, and (2) the presence of highly predictive but uninformative visual features, such as paper color or grain, material signatures and cartographic background textures, or digitization artifacts. Such uninformative features are typically shared among maps with similar production or conservation contexts, regardless of their semantic content or stylistic form. They are treated as confounders, as they may induce overfitting or overconfidence in downstream validation tasks by relying on archival markers that are not directly relevant to cartographic figuration, thereby impairing the interpretability of the representation. The predictive contribution of these features is demonstrated in Experiment 1b (\ref{subsection:exp1b}). It can also be qualitatively observed in Figure \ref{fig:embeddings}, where signs tend to be segregated by background color and where spurious clusters appear due to the presence of encroaching visual objects.

These challenges are addressed by specifically training an adapter module to focus the embedding on relevant visual features. The objective of adaptation is twofold. First, to concentrate the representational power of the encoder on the sign itself while mitigating the impact of encroaching visual objects. Second, to neutralize the impact of paper and background color.

\subsubsection{Foreground segmentation}
\label{subsubsection:foreground_seg}

To focus the representation on signs and diminish the impact of encroaching objects, we implement a lightweight foreground segmentation module. Concretely, a ResNet-18 U-Net model was trained to predict soft foreground probabilities from sign patches using a sigmoid output. Two training datasets were employed for this purpose.

The first training dataset is composed of synthetic training samples generated by overlaying digital icon sprites onto real map images. In total, 679 unique digital icon sprites, created by digital artists for historicized map-art creation, were used. These sprites were overlaid on map crops extracted from the ADHOC database of historical map images \cite{petitpierre_aggregated_2025}, with random size, placement, transparency, and, where applicable, color.

\begin{figure}[t]
\centering
\includegraphics[width=\linewidth]{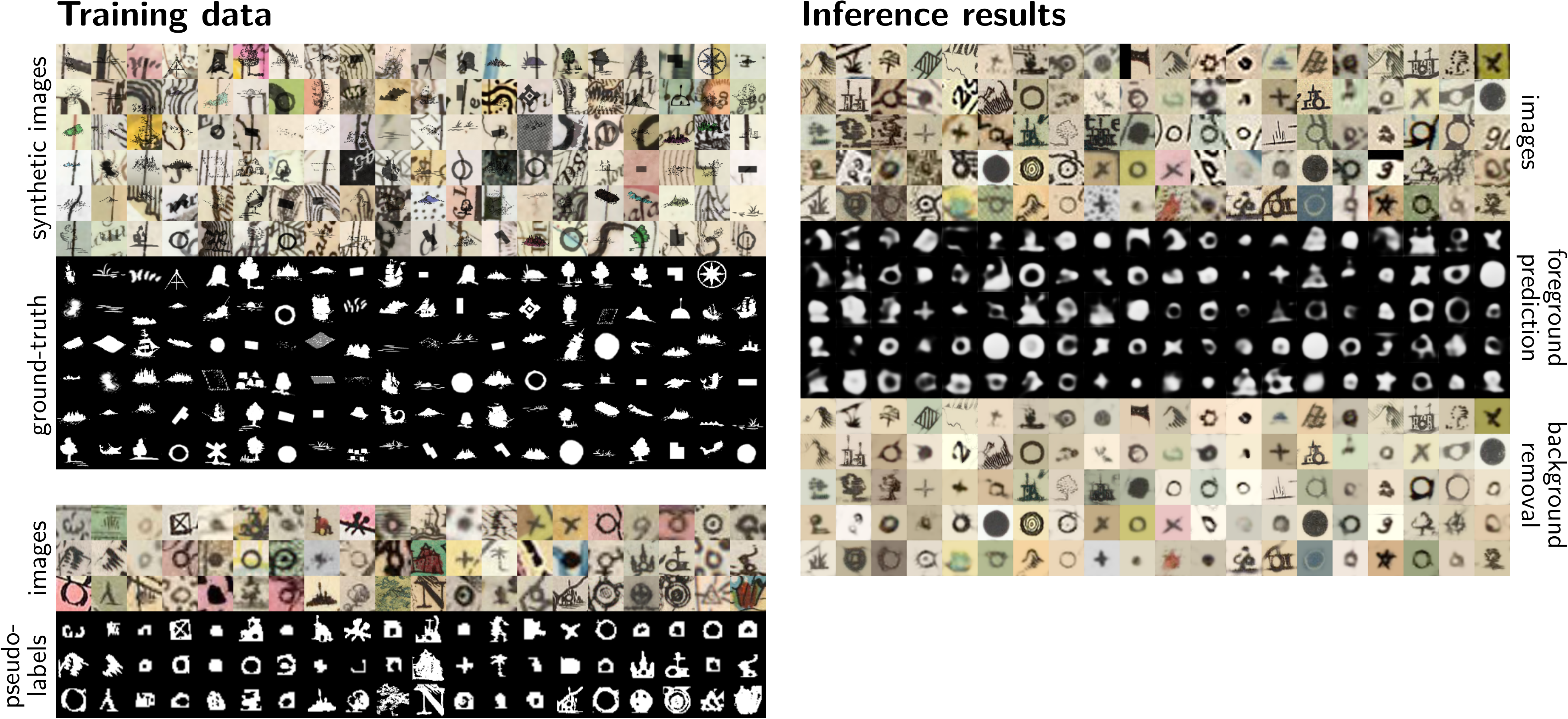}
\caption{\textbf{(left)} Synthetic and pseudolabeled data used for training the foreground segmentation model. Synthetic data are produced by overlaying digital cartographic signs onto map fragments. Pseudolabels are created from real sign samples using traditional computer vision techniques. Synthetic data are associated with perfect masks, while pseudolabels help the model generalize to more realistic sign samples. \textbf{(right)} Inference results. A soft foreground mask is predicted by a neural network, and background content is removed based on the smooth mask. \textit{The foreground segmentation model erases undesirable background content quite effectively.}}
\label{fig:foreground}
\end{figure}

The second training dataset was created by pseudolabeling real sign samples from the CaSiDD dataset. The pseudolabeling approach relied on traditional computer vision heuristics to create approximate foreground masks, employing three simple steps: (1) image binarization, (2) center cropping, and (3) identification of the largest connected component—i.e., the largest graphical feature after binarization.

Individually, both training datasets exhibit strong limitations: low diversity and debatable realism for the first; approximate ground-truth and limited support of edge cases for the second. Arguably, the limitations of each dataset are partially offset by the strengths of its counterpart. In total, 3,800 training samples were created: 2,716 by overlaying icon sprites on real map patches ($71 \%$) and 1,084 by pseudolabeling ($29 \%$). The ResNet-18 U-Net model was then trained for 35 epochs on 3,350 samples drawn with replacement, with a batch size of 50, employing the Adam optimizer \cite{kingma_adam_2015} and mean squared error (MSE) loss. The final model exhibits an MSE of $3 \cdot 10^{-2}$ on the validation set ($n=450$). Representative results of foreground prediction are shown in Figure \ref{fig:foreground}.

The foreground segmentation is an approximate model with its own biases. Rather than expecting it to solve the challenges of sign detection by itself, we use it as a target to learn a softer, adapted representation.

\subsubsection{Adapter}

The adapter module is a block composed of two fully connected (FC) layers with a SiLU \cite{elfwing_sigmoid_2017} activation function ($\mathrm{FC}_{D_{out} \to 384} ,\,\mathrm{SiLU},\, \mathrm{FC}_{384 \to D_{out}}$), where $D_{out}$ is the dimension of the original embedding. The full output is $L_2$-normalized. The adapter is trained with early stopping for up to 4,000 iterations, with a batch size of 128, using extracted map signs from the ADHOC dataset as inputs and their preprocessed counterparts as targets. Preprocessing involves grayscale conversion and smooth background removal using the square roots of the predicted foreground probabilities. With the encoder frozen, training minimizes one minus the mean cosine similarity between full adapted sign embeddings and the frozen, unadapted embeddings of their preprocessed counterparts, using AdamW with an initial learning rate of $10^{-3}$, multiplied by 0.9 every 250 iterations. Semantic and stylometric evaluation uses the normalized 384-dimensional output of the first FC layer, before SiLU.

This adaptation methodology carries explicit design choices, notably the decision to prioritize sign shape over color. This is supported by historical arguments, as color addition was, until the middle of the 19th century, separate from the printing process \cite{verdier_color_2020,woodward_techniques_2007}. As a consequence, color variables generally contribute less to map semantics than shape \cite{petitpierre_generic_2021}; hence the choice to rebalance their impact with respect to the natural images on which VFMs were trained.

\subsection{Experiment 1a: Sign classification (semantics)}
\label{subsection:exp1a}

\subsubsection{Method}
In the first experiment, we assess the expressive power of encoding vectors to embed content and retrieve specific classes of cartographic signs. In this classification task, only the samples labeled with one of the 24 most common semantic classes are retained (i.e., 18,202 out of a total of 18,750). Classification relies on a simple multi-layer perceptron (MLP) with a single 256-unit ReLU hidden layer. To enable comparison, all features have 384 dimensions: wider backbone outputs are reduced using PCA \cite{hotelling1933analysis}, fitted once to the full labelled corpus, while adapted features use the intermediate representation described above.

The samples are partitioned into a training (80 \%) and a validation set (20 \%) using stratified sign-level splits; the features are standardized relative to the training set. The MLP classifier is then optimized on annotated targets. Performance is measured as the class-weighted averages of precision and recall across 100 accepted trials, with weights fixed to the full-corpus class frequencies.

\subsubsection{Results}

\begin{table}[b!]
\caption{Performance metrics for the classification of icons into 24 semantic classes. In each experiment, classification uses the MLP classifier. For both precision and recall, the reported ranges give the 2.5th--97.5th percentiles across 100 trials. The confusion matrix and the detailed per-class performance are provided in Figs. \ref{fig:classification_confusion} and \ref{fig:classification_F1}.}
\label{tab:exp1a}
\begin{tabular}{llll}
Model                      & Treatment                   & Precision {[}95\% range{]}         & Recall {[}95\% range{]}            \\ \hline
\multirow{3}{*}{DINOv2}    & base                        & \textbf{82.5\% {[}81.0–83.9{]}} & \textbf{83.4\% {[}82.1–84.6{]}} \\
                           & grayscale + foreground seg. & 77.3\% {[}76.1–78.7{]}          & 78.4\% {[}77.2–79.7{]}          \\
                           & adapter                     & 81.1\% {[}80.0–82.2{]}          & 81.9\% {[}80.9–83.1{]}          \\ \hline
\multirow{3}{*}{DINOv3}    & base                        & \textbf{79.5\% {[}78.3–80.8{]}} & \textbf{80.6\% {[}79.5–81.7{]}} \\
                           & grayscale + foreground seg. & 74.4\% {[}72.8–75.7{]}          & 76.3\% {[}75.1–77.5{]}          \\
                           & adapter                     & \textbf{78.3\% {[}76.7–79.7{]}} & 79.3\% {[}77.5–80.6{]}          \\ \hline
\multirow{3}{*}{RADIOv3} & base                        & \textbf{83.1\% {[}82.1–84.2{]}} & \textbf{83.8\% {[}83.1–84.7{]}} \\
                           & grayscale + foreground seg. & 77.3\% {[}76.2–78.3{]}          & 78.8\% {[}77.8–79.6{]}          \\
                           & adapter                     & \textbf{82.4\% {[}80.8–83.7{]}} & \textbf{83.1\% {[}81.7–84.3{]}} \\ \hline
\multirow{3}{*}{RADIOv4} & base                        & 84.4\% {[}83.1–85.4{]}          & 84.9\% {[}83.7–86.0{]}          \\
                           & grayscale + foreground seg. & 77.2\% {[}76.0–78.4{]}          & 78.4\% {[}77.1–79.6{]}          \\
                           & adapter                     & \textbf{87.0\% {[}85.7–88.2{]}} & \textbf{87.4\% {[}86.1–88.6{]}}
\end{tabular}
\end{table}

\begin{figure}[h!]
\centering
\includegraphics[width=\linewidth-100pt]{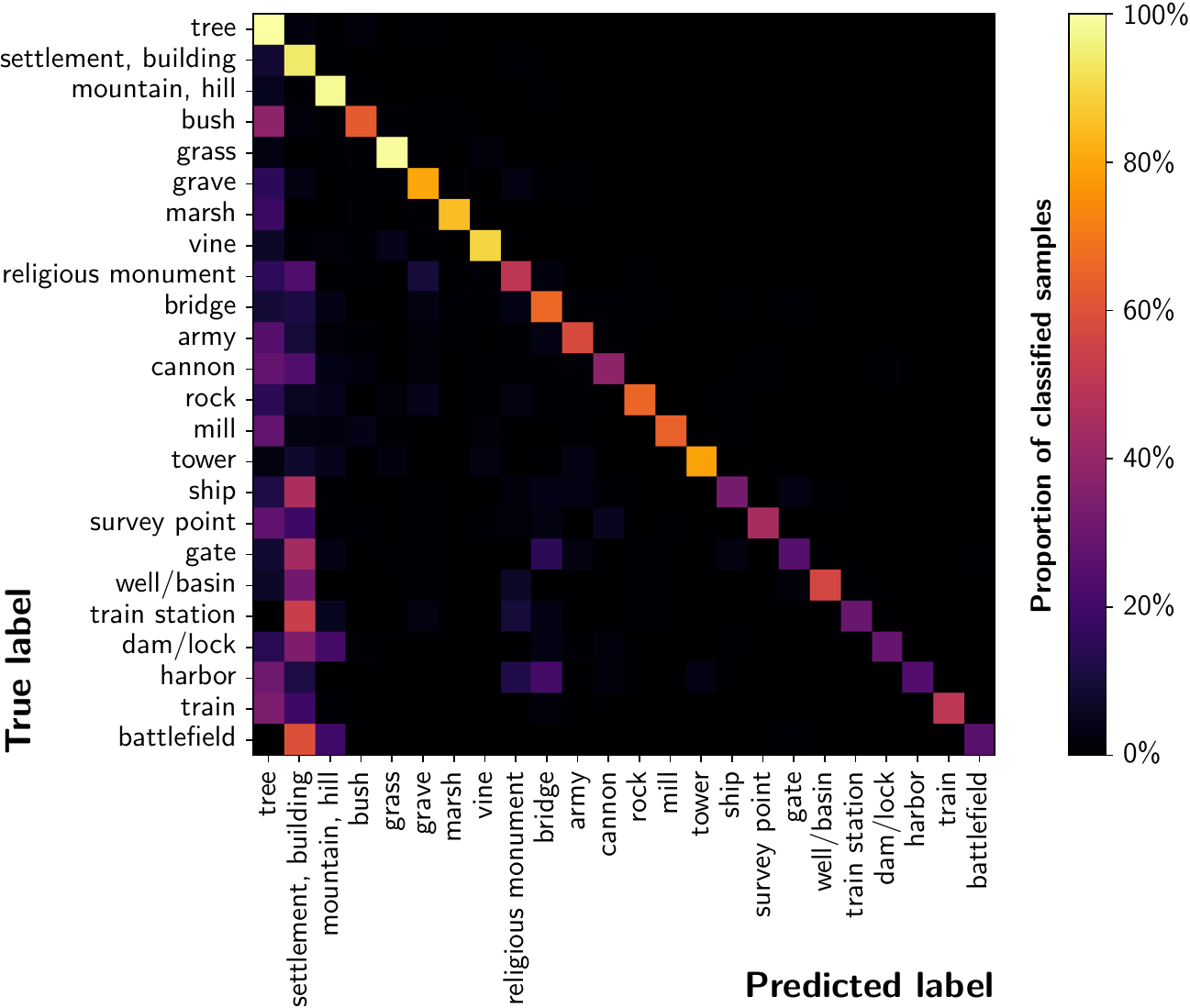}
\caption{Confusion matrix for the sign classification task, computed from the embeddings generated by the adapted RADIOv4, using the MLP classifier and 24 classes. The reported results aggregate 100 trials, each with $n_{val} = 3,641$ validation signs. Color intensity denotes the mean raw confusion matrix normalized by true-label row. The diagonal entries therefore represent recall. Labels are ordered by their frequency in the training set (from $n_1=6,679$ for trees to $n_{24}=4$ for battlefield). \textit{For example, bushes are frequently mistaken for trees.}}
\label{fig:classification_confusion}
\end{figure}

\begin{figure}[h!]
\centering
\includegraphics[width=\linewidth]{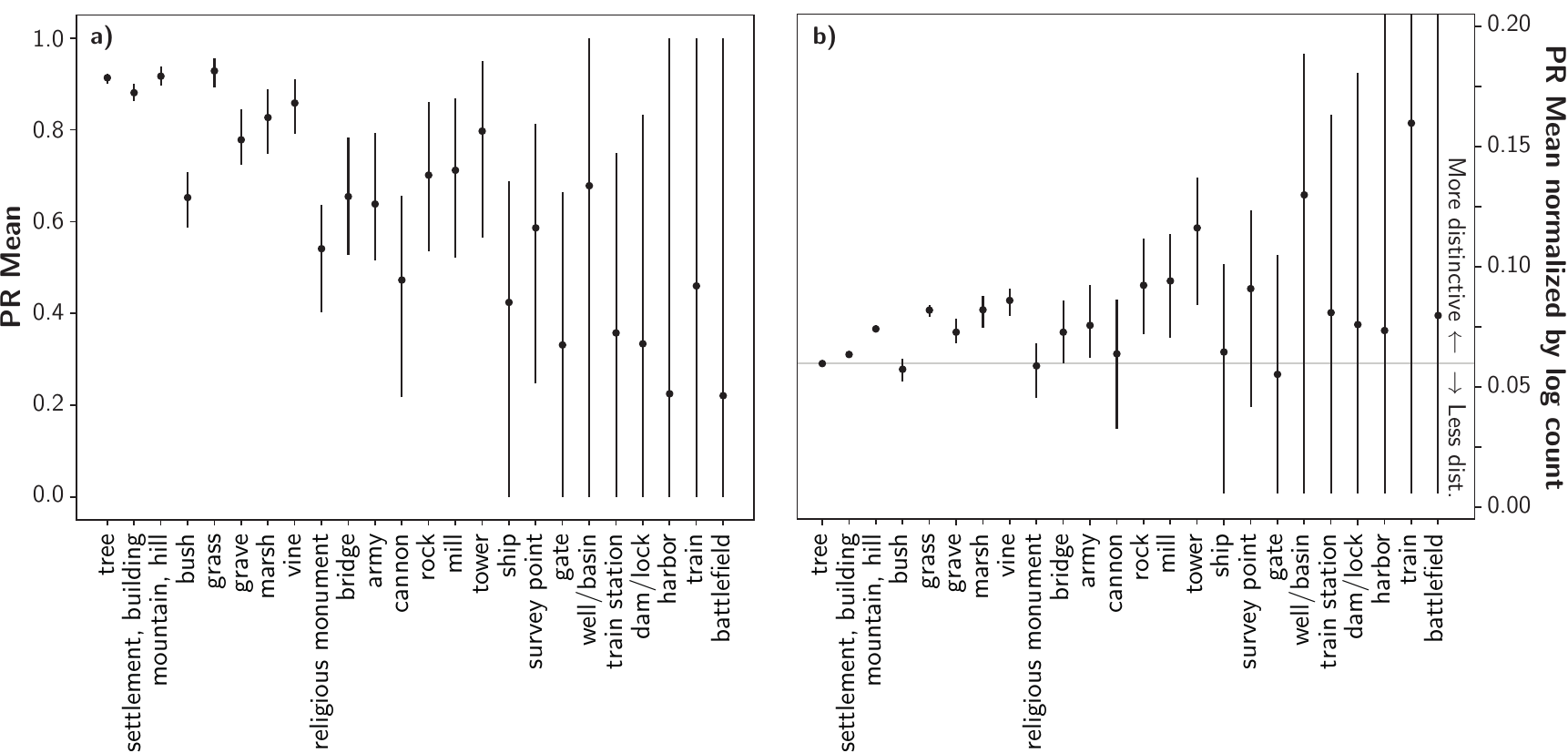}
\caption{Precision-Recall (PR) mean of the sign classification task for each semantic class, computed from the embeddings generated by the adapted RADIOv4, using the MLP classifier and 24 classes. The reported results aggregate 100 trials, each with $n_{val} = 3,641$ validation signs. The vertical bars denote the 2.5th--97.5th percentiles across 100 trials. Labels are ordered by their frequency in the training set (from $n_1=6,679$ for trees to $n_{24}=4$ for battlefield). The mean precision and recall values are reported in Table \ref{tab:exp1a}. Plot \textbf{(a)} reports the original PR-Mean for each class, where PR-Mean = (Precision + Recall) / 2, whereas subfigure \textbf{(b)} presents the PR-Mean divided by the natural logarithm of the training set class counts. The horizontal bar indicates the mean of the tree class. \textit{For example, grass and vine signs are classified well, especially considering their frequency in the training data.}}
\label{fig:classification_F1}
\end{figure}

The results of the classification experiment are reported in Table \ref{tab:exp1a}. The base and adapted models perform similarly, although the highest precision and recall scores are achieved using an adapted RADIOv4 backbone (precision $=87.0\%$ [95\% trial range: 85.7--88.2\%], recall $=87.4\%$ [95\% trial range: 86.1--88.6\%]). Preprocessing appears to negatively impact classification performance.

However, we consider two distinct notions of success: on the one hand, the predictive accuracy; on the other, the interpretive validity. The generally high performance of the base model may indeed be partly overstated, as it could be driven by the reliance on material confounders, that is, highly predictive features related to digitization and support characteristics (see Exp. 1b, \ref{subsection:exp1b}).

Whereas preprocessing through grayscale conversion and foreground segmentation tends to attenuate the spurious contribution of weak visual features (see Exp. 3, \ref{subsection:exp3}), it may also degrade image quality. The adapted embeddings, by contrast, exhibit a classification performance similar to, and sometimes higher than, the base setting, while the full adapter outputs show stronger associations with shape and weaker associations with background color (cf. \ref{subsection:exp3}). From this perspective, adapted embeddings are preferable to no treatment.

Figure \ref{fig:classification_F1} reports the mean precision-recall score breakdown per class. In most cases, the classification accuracy approximately depends on the log count of labeled instances. One tendentially underperforming class is the bush class, which is often confused with trees, as visible in Figure \ref{fig:classification_confusion}. Tree-settlement and tree-bush confusions are prominent errors in this experiment. Other common confusions include the building/settlement class and specific built infrastructures, like dams, locks, gates, wells, and religious monuments, as well as boats and harbors. In most cases, confusion can be attributed to a degree of semantic and figurative proximity between these classes, a similar or uncharacteristic context of occurrence, or to data imbalance.

\subsection{Experiment 1b: Impact of paper and background color on classification}
\label{subsection:exp1b}

\subsubsection{Method}
The aim of this complementary experiment is to demonstrate the opportunity for weak features, related to the grain and color of the map material (e.g., paper type) and to the specifics of digitization (e.g., scanner), to bias classification, as hypothesized above. For this purpose, the same labelled corpus is evaluated using simple RGB color descriptors targeting the sample background. The descriptors use weights $1-\sqrt{p}$, where $p$ is the foreground probability predicted by the segmentation module (\ref{subsubsection:foreground_seg}). A logistic regression model, with training-fitted feature standardization, is then used to categorize the samples into 24 classes. The experiment is repeated 100 times, assigning each sign independently to training with probability 0.8 and otherwise to validation.

\subsubsection{Results}
Table \ref{tab:exp1b} shows the results of the experiment. As expected, the RGB color descriptors are less discriminative than the features derived from the VFMs (Table \ref{tab:exp1a}). However, their precision is higher than that of a majority-class baseline that would always predicts trees. This indicates that the logistic regression model can learn predictive information from these weak features. In other words, background color can influence classification through cues unrelated to sign semantics. This result is consistent with the high precision observed for the base model in Experiment 1a (\ref{subsection:exp1a}), and globally supports the use of preprocessing or adaptation strategies to improve generalizability and feature interpretability.

\begin{table}[h]
\caption{Impact of weak features on sign classification. Classification uses logistic regression. Scores use fixed full-corpus weighting by class frequency. For both precision and recall, the ranges give the 2.5th--97.5th trial percentiles. The last row is the constant tree class baseline.}
\label{tab:exp1b}
\begin{tabular}{lll}
Treatment                  & Precision {[}95\% range{]} & Recall {[}95\% range{]}   \\ \hline
color background (base)                        & 38.8\% {[}36.2–40.1{]}  & 51.6\% {[}51.1–52.0{]} \\
\emph{tree class baseline}                     & \emph{25.6\% {[}24.9–26.2{]}}  & \emph{50.6\% {[}50.6–50.6{]}}
\end{tabular}
\end{table}

\subsection{Experiment 2: Stylometry and attribution}

The aim of the second experiment is to assess the ability of the embeddings to identify replicated signs and to encode cultural or stylistic similarities. Here, we consider the relationship between encoding distances and homogeneous communities of mapmakers identified in the ADHOC dataset. Maps created by the same group of makers, or closely related creators—working for instance, in the same cartographic workshop—should exhibit similar sign forms, share a common repertoire and material features. Here, we aim to disentangle the former two dimensions from the latter.

\subsubsection{Definitions}

Let \( D \) be the dataset of maps, produced by mapmakers belonging to a finite set of communities \( C \). For each community \( c \in C \), let \( M_c \subset D \) denote the set of maps created by that community:
\begin{equation}
D = \bigcup_{c \in C} M_c .
\end{equation}
We write \( c(m) \in C \) for the community responsible for a map \( m \). Each map \( m \) is considered a collection of signs
\begin{equation}
X_m = \{ x_m^1, x_m^2, \dots, x_m^{N_m} \},
\end{equation}
where $N_m \leq 128$ is the number of extracted icons. Each icon \( x \in X_m \) is embedded as a vector $z \in \mathbb{R}^d$, where \( d \) denotes the dimension of the embedding space (here, $d=384$). All embeddings are $L_2$-normalized, which is consistent with the subsequent computation of angular similarities.

\subsubsection{Kernel model of sign embeddings}

Each map is represented by an unnormalized mixture of kernels over its sign embeddings. Given a sign $x$ and a map $m$ with sign embeddings
\(
\{z_m^1, \dots, z_m^{N_m}\},
\)
the kernel score of \( z \) under map \( m \) is defined as
\begin{equation}
q\!\left(z \mid m\right)
=
(1 - \alpha)\,
\frac{1}{N_m}
\sum_{i=1}^{N_m}
\exp\!\left(
-\frac{\lVert z - z_m^i \rVert^2}{2h^2}
\right)
+
\alpha\, b_0\!\left(z\right),
\end{equation}
where $h$ is the median Euclidean distance over up to 20,000 sampled sign pairs within the batch, and $\alpha$ is a small stabilizing coefficient ($\alpha=10^{-3}$). The base kernel is
\(
b_0\!\left(z\right)
=
\exp\!\left(-\frac{\lVert z-\mu \rVert^2}{2\sigma^2}\right),
\)
with \( \mu \) the mean embedding and \( \sigma \) the mean coordinate-wise sample standard deviation, estimated from all selected sign embeddings in each batch. Scores are computed in log space for numerical stability.

\subsubsection{Map-to-map semiotic dissimilarity}

Given two maps $m$ and $m'$, the \emph{directed} stylometric similarity $s(m \mid m')$ is defined as the mean log kernel score of the signs of $m$ under the kernel model of $m'$. Intuitively, the map $m$ is considered similar to $m'$ if its signs tend to be close in the representation space to the signs in $m'$. Contrary to optimal transport problems like the Earth Mover's Distance, this does not require mapping each sign in $m$ to a sign in $m'$, making the measure less sensitive to relative sign frequencies.

\begin{equation}
s(m \mid m') = \frac{1}{N_m} \sum_{i=1}^{N_m} \log q\!\left(z_m^i \mid m'\right)
\end{equation}

Generalizing, the symmetric dissimilarity $\delta_{m, m'}$ between the maps \( m \) and \( m' \) can then be computed as
\begin{equation}
\label{eq:map-to-map}
\delta_{m, m'}
=
\max\!\left\{0,\,
\frac{1}{2}
\Bigl[
s(m \mid m) - s(m \mid m')
+
s(m' \mid m') - s(m' \mid m)
\Bigr]\right\}.
\end{equation}

Self-normalization ($s(m \mid m)$ and $s(m' \mid m')$ terms) compensates for the inherently lower pairwise similarity of internally diverse maps; these self scores include each sign's own kernel.

\subsubsection{Evaluation of community clustering tendency}

The 848 communities are traversed in 85 consecutive batches: 84 groups of ten communities and a final group of eight. Each batch draws up to five map records per community and up to 128 signs per map, without replacement; missing or empty map files are skipped. For each batch \( B = \{m_1, \dots, m_n\}, n \leq 50 \), the symmetric matrix of pairwise map dissimilarities is calculated as given in Equation \ref{eq:map-to-map}.

Clustering tendency is evaluated using (i) the ROC-AUC separating within- and between-community pairs, with within-community pairs positive and the negative dissimilarity as the score, and (ii) the within-community matching rate (wCM@k), corresponding to the average proportion of the $k$ nearest neighbors sharing the same community label. Metrics are averaged over the 85 community batches.

\subsubsection{Results}

\begin{table}[b]
\caption{Performance of the stylometric experiment in each setting. ROC-AUC measures the contrast between within- and between-community pairs. wCM@k indicates the within-community matching rate, i.e., the average proportion of nearest neighbors sharing the same community label for any item. CI denotes the Student-$t$ 95\% confidence interval of the mean across 85 community batches.}
\label{tab:exp2}
\begin{tabular}{lllll}
Model                      & Treatment                   & ROC-AUC {[}95\% CI{]}               & wCM@1 {[}95\% CI{]}             & wCM@3 {[}95\% CI{]}             \\ \hline
\multirow{3}{*}{DINOv2}    & base                        & \textbf{74.9\% {[}73.3–76.4{]}} & \textbf{63.2\% {[}60.3–66.2{]}} & \textbf{50.2\% {[}47.3–53.1{]}} \\
                           & grayscale + foreground seg. & 73.6\% {[}72.1–75.1{]}          & 59.1\% {[}56.0–62.2{]}          & 45.9\% {[}43.1–48.8{]}          \\
                           & adapter                     & \textbf{75.3\% {[}73.7–76.8{]}} & 59.1\% {[}56.1–62.1{]}          & 45.9\% {[}43.1–48.7{]}          \\ \hline
\multirow{3}{*}{DINOv3}    & base                        & \textbf{75.6\% {[}74.0–77.3{]}} & \textbf{65.9\% {[}63.0–68.7{]}} & \textbf{52.2\% {[}49.1–55.2{]}} \\
                           & grayscale + foreground seg. & 74.1\% {[}72.5–75.7{]}          & 60.6\% {[}57.5–63.7{]}          & 48.0\% {[}45.1–50.9{]}          \\
                           & adapter                     & \textbf{76.5\% {[}74.9–78.2{]}} & 62.5\% {[}59.5–65.5{]}          & \textbf{49.7\% {[}46.7–52.7{]}} \\ \hline
\multirow{3}{*}{RADIOv3} & base                        & 71.2\% {[}69.6–72.8{]}          & \textbf{58.1\% {[}54.8–61.4{]}} & \textbf{44.2\% {[}41.3–47.1{]}} \\
                           & grayscale + foreground seg. & 70.4\% {[}69.0–71.8{]}          & 52.9\% {[}50.2–55.6{]}          & 40.0\% {[}37.6–42.4{]}          \\
                           & adapter                     & \textbf{73.7\% {[}72.2–75.2{]}} & \textbf{55.9\% {[}52.5–59.3{]}} & \textbf{43.5\% {[}40.6–46.3{]}} \\ \hline
\multirow{3}{*}{RADIOv4} & base                        & \textbf{77.4\% {[}75.8–79.1{]}} & \textbf{68.3\% {[}65.6–71.1{]}} & \textbf{55.0\% {[}52.0–58.0{]}} \\
                           & grayscale + foreground seg. & 76.4\% {[}74.8–78.0{]}          & 63.5\% {[}60.5–66.5{]}          & 50.7\% {[}47.7–53.7{]}          \\
                           & adapter                     & 75.4\% {[}74.0–76.9{]}          & 63.8\% {[}60.9–66.8{]}          & 49.6\% {[}46.9–52.4{]}         
\end{tabular}
\end{table}

Table \ref{tab:exp2} shows complex evidence regarding the ability of the adapter to improve workshop attribution predictions. As in Experiment 1a (\ref{subsection:exp1a}), the preprocessing involving grayscale conversion and foreground segmentation often degrades performance, likely due to the loss of visual information. Compared with the base models, adaptation yields higher ROC-AUC except for RADIOv4, but lower within-community matching rates for all four models.

These results may initially seem to contrast with those of the first experiment (\ref{subsection:exp1a}), but they are consistent with the interpretation of the latter and with the results of experiments 1b (\ref{subsection:exp1b}) and 3 (\ref{subsection:exp3}). Whereas the removal of confounding archival features has only a moderate effect on sign categorization, it seems intuitively more strongly correlated with the objective of community matching, which is essentially associated with material and digitization factors through production and collection contexts. As such, the drop in within-community matching rates caused by the adapter may be attributable to the disappearance of highly predictive, but also sensitive, material cues. This interpretation is also supported by the larger drop in performance observed in wCM@k compared to ROC-AUC, which suggests that the local neighborhood–plausibly relying on non-structural, weak graphical features–is disrupted first.

Overall, although the final performance metrics (ROC-AUC = 75.4\%, wCM@1 = 63.8\%, wCM@3 = 49.6\%) do not point to complete separability of mapmaker communities according to sign usage, they nevertheless imply significant contrasts. Thus, sign form appears a potent, albeit not entirely discriminative, stylometric cue in cartography.

\subsection{Experiment 3: Feature interpretation and explainability}
\label{subsection:exp3}

The third experiment aims to measure the interpretability, or explainability, of the latent representation. It also attempts to make explicit the impact of preprocessing and adaptation on the association with specific visual cues. In this framework, interpretability is operationalized as the alignment between distances in the embedding space and explicit visual variables (color, shape), related to Bertin's \emph{Semiology of Graphics} \cite{bertin_semiologie_1967}.

\subsubsection{Method}

\begin{table}[b!]
\caption{Ability of the embeddings to describe graphical differences in RGB color and shape. Each value indicates the linear association between distances in the vector spaces, measured on $8 \cdot 10^6$ sampled distance comparisons using Pearson’s coefficient. The best scores are highlighted in bold. For background color, lower values are preferable (↓).}
\label{tab:exp3}
\begin{tabular}{lllll}
Model                      & Treatment                   & Background color ↓ & Foreground color ↑ & Shape (HOG) ↑    \\ \hline
\multirow{3}{*}{DINOv2}    & base                        & 15.9\%                 & 7.9\%                  & 23.5\%          \\
                           & grayscale + foreground seg. & 8.7\%                  & \textbf{8.5\%}         & \textbf{32.3\%} \\
                           & adapter                     & \textbf{5.2\%}         & 4.4\%                  & 30.9\%          \\ \hline
\multirow{3}{*}{DINOv3}    & base                        & 17.5\%                 & 15.5\%                 & 9.8\%           \\
                           & grayscale + foreground seg. & 12.2\%                 & 13.7\%                 & \textbf{20.2\%} \\
                           & adapter                     & \textbf{11.9\%}        & \textbf{17.8\%}        & 19.2\%          \\ \hline
\multirow{3}{*}{RADIOv3} & base                        & 12.8\%                 & \textbf{9.0\%}         & 5.3\%           \\
                           & grayscale + foreground seg. & 3.2\%                  & 7.9\%                  & 11.7\%          \\
                           & adapter                     & \textbf{0.8\%}         & 5.4\%                  & \textbf{32.8\%} \\ \hline
\multirow{3}{*}{RADIOv4} & base                        & 11.2\%                 & 7.0\%                  & 27.8\%          \\
                           & grayscale + foreground seg. & 7.9\%                  & 7.5\%                  & 31.7\%          \\
                           & adapter                     & \textbf{6.8\%}         & \textbf{9.6\%}         & \textbf{39.2\%}
\end{tabular}
\end{table}

In this setting, the pairwise distances within each embedding space were compared to the distances obtained from baseline computer vision descriptors targeting color (RGB) and sign shape, that is morphology, respectively. Intuitively, this corresponds to measuring the ability of the representation spaces to express meaningful visual information.

The color space is represented by the red-green-blue (RGB) channels of the digital image. Average foreground and background colors are measured as weighted mean RGB values, using $w=\sqrt{p}$ and $1-w$, respectively, where $p$ is the predicted foreground probability and $w$ is set to zero for black pixels.

Morphological descriptors are obtained by computing the histogram of oriented gradients (HOG, \cite{mcconnell_method_1986}). The key idea behind the HOG algorithm is that shapes can be effectively captured by the distribution of edge orientations. In the present experiment, signs are resized to $48 \times 48$ pixels and cropped to $44 \times 44$ pixels. Gradients are computed over $9 \times 9$ pixel cells, using nine orientation bins and blocks of $3 \times 3$ cells, resulting in 729-dimensional shape descriptors.

Finally, a sample of 4,000 signs is randomly drawn from distinct maps. Icons are embedded using DINO or RADIO encoders, with or without adaptation and preprocessing. In parallel, RGB and HOG descriptors are computed on the same sign samples. Distances are then calculated for 8 million comparisons using column sampling with replacement followed by lower-triangular selection, with cosine distance for features derived from VFMs or HOG and Euclidean distance for RGB descriptors. Structural similarity between the representation spaces is defined as the linear relationship between paired distances, quantified by the Pearson coefficient.

\subsubsection{Results}

Table \ref{tab:exp3} summarizes the results of the feature explanation experiment. Both preprocessing and adaptation reduce the association of embedding distances with background color. By contrast, they also seem to effectively focus the embedding on the image foreground, i.e., the sign, with shape, and in some cases color, becoming more prominent. DINO models generally seem to focus more on color compared to RADIO models.

RADIOv4 is characterized by its superior expressiveness for sign shape, both with and without an adapter, which may explain, in part, its good performance in both classification and community-contrasting tasks.

\section{General Discussion and Conclusion}

This paper presented and validated a methodology for the scalable, generic detection of map signs, and the expressive encoding of both geographic content and stylistic, author-related features. The final representation, based on an adapted RADIOv4, is interpretable and specific, targeting sign shape and color.

The experiment on the extraction of cartographic signs highlighted the limits of the ontological assumptions of object detection frameworks in this particular context: spatial boundedness, discreteness, and minimal overlap. The arbitrary character inherent to sign labels is particularly reflected in the rather low recall rate, although a qualitative examination qualifies these results. In Figure \ref{fig:detection_results}, false positive errors appear especially rare. Owing to the low recall, the performance is insufficient for the exhaustive reconstruction of past landscapes. However, the detector may support comparative and distributional analyses, such as estimating relative frequencies and spatial co-occurrence patterns. Consistently, it yields substantial potential for the geographic history of cartography, enabling the investigation of semiotic culture at scale, and the spatial dynamics of sign diffusion \cite{petitpierre_studying_2025}.

Map sign encoding was primarily achieved using a neural adapter trained under domain-specific constraints, including the challenge of material confounders. Pragmatically, the adapter is effective in producing expressive, interpretable features for downstream classification or stylometric tasks, compared to generic VFM backbones. The evaluation was driven by the notion that not only raw performance gains count, but also the expressiveness of the representation and its interpretability. Computationally, an adapter head is efficient compared to fine-tuning the VFM. It also prevents the risk of catastrophic forgetting. At inference time, it is also significantly lighter than the ResNet-based preprocessing, decreasing the average sign embedding time from 55 ms to 16 ms.

Computational efficiency in detection and encoding also opens the way for integrating these models into interactive research interfaces. Beyond retrieval, the detection of map signs at scale concretely enables new lines of research. In cartographic semiotics, history, and cultural science, it enables tracking the diffusion of icons and symbols across regions and historical periods, quantifying the impact of technical disruptions on figurative shifts, and empirically testing prototype and exemplar theories against observed data. In historical geography and related fields, it could for instance indicate the location and importance of settlements before the start of censuses in colonized territories, document the evolution of vineyard extent after epiphytotic events, or enable the density mapping of historic peatlands with high potential for carbon storage.

\begin{acknowledgments}

\paragraph{Thanks.} We wish to acknowledge the contribution of Jiaming Jiang for annotating a portion of the CaSiDD dataset.

\paragraph{Funding.} This work was funded by the public endowment of EPFL, Swiss Federal Institute of Technology in Lausanne. It has also received funding from the European Union's Horizon Europe research and innovation programme under grant agreement No. 101233051.

\paragraph{Data availability statement}
The sign detection training dataset and the ADHOC database used for attribution in the second experiment are available open access on Zenodo \cite{petitpierre_cartographic_2025,petitpierre_aggregated_2025}.

\end{acknowledgments}

\section*{Declaration on Generative AI}
During the preparation of this work, the author used GPT-5.2 and DeepL's translation models for paraphrasing and rewording, writing style improvements, grammar and spelling checks, and text translation. Conversely, AI was not employed for the production of scientific content, such as the retrieval of scientific references, the description of the methodology, the analysis or the discussion of results, or the formulation of conclusions. After using these tools, the author reviewed and edited the content as needed and takes full responsibility for the publication’s content. 

\begingroup
\interlinepenalty=10000
\bibliography{bibliography}
\endgroup

\end{document}